\documentclass{article}
\usepackage[utf8]{inputenc}
\usepackage{authblk}
\usepackage{changepage} 
\usepackage{natbib}
\usepackage{graphicx}
\usepackage[hang,flushmargin]{footmisc} 
\usepackage{amssymb}
\usepackage{amsmath}

\usepackage{natbib}
\setcitestyle{authoryear,open={(},close={)}}

\usepackage{xcolor}
\usepackage{chemformula}
\usepackage{booktabs}
\usepackage{multirow}

\usepackage{titlesec}

\titleformat{\paragraph}
{\normalfont\normalsize\bfseries}{\theparagraph}{1em}{}
\titlespacing*{\paragraph}
{0pt}{3.25ex plus 1ex minus .2ex}{1.5ex plus .2ex}

\title{On the attribution of confidence \\ to large language models}

\author{Geoff Keeling and Winnie Street}

\affil{Google Research}

\begin{document}

\maketitle

\begin{abstract}
     
     Credences are mental states corresponding to degrees of confidence in propositions. Attribution of credences to Large Language Models (LLMs) is commonplace in the empirical literature on LLM evaluation. Yet the theoretical basis for LLM credence attribution is unclear. We defend three claims. First, our \textit{semantic} claim is that LLM credence attributions are (at least in general) correctly interpreted literally, as expressing truth-apt beliefs on the part of scientists that purport to describe facts about LLM credences. Second, our \textit{metaphysical} claim is that the existence of LLM credences is at least plausible, although current evidence is inconclusive. Third, our \textit{epistemic} claim is that LLM credence attributions made in the empirical literature on LLM evaluation are subject to non-trivial sceptical concerns. It is a distinct possibility that \textit{even if} LLMs have credences, LLM credence attributions are generally false because the experimental techniques used to assess LLM credences are not truth-tracking.

\end{abstract}

\section{Introduction}

Large Language Models (LLMs) such as GPT-4 \citep{OpenAI2023gpt} and Gemini \citep{team2023gemini} are neural networks trained to predict the next word in a sequence of words on massive text datasets.\footnote{We are grateful to Benjamin Lange, Michael Graziano, Nino Scherrer, John Oliver Siy, and Blaise Ag\"uera y Arcas for comments and critical discussions, and to audiences at Princeton University, Hamburg University of Technology, and Google Research.} Training for next word prediction on broad data allows LLMs to acquire capabilities including theory of mind, spatial reasoning, and coding \citep{bubeck2023sparks}. LLMs have given rise to a vast empirical literature that seeks to identify and evaluate their capabilities.\footnote{\citet{chang2023survey} give a survey of the LLM evaluation literature. Notable contributions to LLM evaluation include BIG-Bench \citep{srivastava2022beyond} and HELM \citep{liang2022holistic}.}

Scientists engaged in LLM evaluation often attribute \textit{credences} to LLMs; that is, degrees of belief or confidence in propositions. Credences are attributed to LLMs on the basis of experimental techniques such as prompting the LLM to report a credence or examining the distribution of answers given by the LLM in response to a question over several independent trials \citep{geng2023survey}. Some scientists claim that prominent LLM evaluation benchmarks such as the HuggingFace Open LLM Leaderboard ‘posess a significant limitation' in failing to register that ‘two LLMs may achieve identical accuracy scores but exhibit different levels of uncertainty regarding the question' \citep[1-2]{ye2024benchmarking}.

LLM credence attribution is puzzling. Credences are normally understood as psychologically real mental states, attributable to humans, that alongside utilities causally explain behaviour \citep[c.f.][]{okasha2016interpretation, dietrich2016mentalism}.\footnote{This interpretation of credences is a \textit{mentalist} interpretation on which credences and utilities are quantitative extensions of belief-desire psychology. Credences can also be framed along behaviourist lines as formal constructs derivable from revealed preferences, as is common in microeconomics \citep{von2007theory, savage1972foundations}. See \S \ref{Metaphysical Anti-Realism} for discussion.} Attribution of credences to LLMs -- which do not obviously have mental states and which cannot obviously perform acts that are causally explainable with reference to mental states -- requires philosophical analysis. The obvious concern here is that LLM credence attribution may turn out to be \textit{complete bunkum}. If this is the case, then we must re-evaluate the implications that have been drawn from LLM credence attributions for practically and ethically significant LLM characteristics, including factuality, conceptual understanding, and honesty.     

We explore three questions. First, are scientists \textit{literally} ascribing degrees of confidence to LLMs, or are LLM credence attributions correctly interpreted as a stylised shorthand for non-mentalistic claims about LLMs? We argue that LLM credence attributions are (at least in general) correctly interpreted literally as expressing truth-apt beliefs on the part of scientists that purport to describe facts about LLM credences. Second, do LLM credences exist? We argue that the existence of LLM credences is at least plausible, but current evidence is inconclusive. Third, supposing that LLMs have credences, are current techniques for assessing LLM credences reliable? We argue that current techniques for assessing LLM credences are subject to non-trivial sceptical worries. It is a distinct possibility that \textit{even if} LLMs have credences, LLM credence attributions made on the basis of existing experimental techniques are generally false.

In \S 2, we introduce LLMs. In \S 3, we characterise the scientific context in which LLM credence attributions are made, and distinguish between three empirical techniques for assessing LLM credences. In \S 4, we discuss the semantic interpretation of LLM credence attributions. In \S 5, we discuss the existence of LLM credences. In \S 6, we discuss the epistemic status of LLM credence attributions given existing empirical assessment techniques. In \S 7, we conclude.

\section{Large Language Models}

Language models are statistical models that, roughly, predict the next word given a sequence of words. Specifically, language models are functions. The inputs are sequences of \textit{tokens}, which are syntactic units that include words, sub-words, and punctuation symbols. The outputs are probability distributions over tokens in the model's vocabulary such that the probability assigned to each token is an estimate of how likely each token is to succeed the input sequence.\footnote{In practice, LLMs use log-probabilities instead of probabilities. There are good practical reasons for using log-probabilities in computational contexts. For example, consider a sequence of 50 words each with a probability of 0.8. The overall probability of the sequence is $0.8^{50}$ = $0.00001427247$. Small probabilities like these can easily underflow to $0$ in floating-point arithmetic. Log-probabilities have a greater dynamic range and so avoid underflow issues; in this case, for example, $\ln(0.00001427247)$ $=$ $-11.1571780511$. Nothing of importance hinges on whether we talk about log-probabilities or probabilities for the purposes of this paper.} 

Large Language Models (LLMs) are neural networks that \textit{instantiate} language models. The inputs to the network are arrays of numbers which represent token sequences. Each token in the model's vocabulary is assigned a numerical ID such that any given token sequence can be represented as an array of numbers. Second, for each input sequence, the network outputs an array of real-numbers which are each between $0$ and $1$ and which sum to $1$. These numbers are probability estimates corresponding to each token in the LLM's vocabulary.

How do LLMs compute token succession probabilities? First, the input sequence is passed through an \textit{embedding layer}, which maps each token ID in the sequence to a vector (or point) in a high-dimensional space. Each vector is called a \textit{token embedding}. Roughly, tokens that tend to show up in similar linguistic contexts map to embeddings that are closer together in the embedding space. Second, the embeddings for each token in the input sequence are fed into the main body of the neural network -- the \textit{transformer} \citep{vaswani2017attention}. Transformers contain a sequence of blocks. Each block generates an embedding for every token in the input sequence which represents the linguistic relationships that each token stands in to every other token in the input sequence. The embeddings generated by block $N$ are inputs to block $N+1$, such that each successive block generates a more refined picture of how each token in the input sequence relates to every other token. The output of the final block is used to compute probabilities for each token in the LLM's vocabulary. In particular, the embeddings outputted by the final block are passed through a \textit{linear layer} which projects them into $K$-dimensional space, where $K$ is the number of tokens in the model's vocabulary. The linear layer outputs a set of real-valued \textit{logits} for each token, where higher logit values indicate that the relevant token is more likely to succeed the input sequence. The logits are normalised into probabilities (real-numbers between 0 and 1 that sum to 1) in the final layer of the network.

LLMs are developed through a training process. At a high-level, the neural network contains hundreds of billions of adjustable parameters. Each set of parameter values corresponds to a particular algorithm for computing token succession probabilities from token sequences. In training, the model predicts masked tokens from preceding tokens using internet text data. The probability distribution outputted by the LLM conditional on a given input sequence is evaluated against a \textit{loss function} which penalises the LLM in proportion to the distance between the LLM's probability distribution and the degenerate distribution which assigns probability 1 to the true successor token and 0 to all other tokens. An algorithm computes the gradients of the loss with respect to the model parameters, and updates the parameters in the opposite direction of the gradient so as to reduce the model's loss on each particular example. This ‘hill climbing' process is iterated until the model's probability estimates closely approximate the true probability distribution of tokens in the training data. 

The critical insight is that in learning to predict the next token successfully on vast quantities of internet text data, LLMs acquire useful capabilities that can be framed as next token prediction problems. This includes general knowledge (\texttt{Paris is the capital of\rule{0.5cm}{0.15mm}}), specialist knowledge (\texttt{\ch{H2SO4} + \ch{2 NaOH} $\rightarrow$ \ch{Na2SO4} +\rule{0.5cm}{0.15mm}}), coding (\texttt{$\text{W}_{\text{key}}$ = torch.nn.Parameter(torch.\rule{0.5cm}{0.15mm}}), logical reasoning (\texttt{If $A$ $\rightarrow$ $B$ and $B$ $\rightarrow$ $C$, then $A$ $\rightarrow$\rule{0.5cm}{0.15mm}}), and spatial reasoning (\texttt{$A$ is behind $B$, so $B$ is in front of\rule{0.5cm}{0.15mm}}) \citep{bubeck2023sparks}.

Last, sampling from language models can generate text. Given the input \texttt{Who directed Fight Club?}, the model returns a probability distribution over tokens. Sampling from this distribution (that is, roughly, selecting a token via a weighted lottery where the weights are given by token probabilities) may return \texttt{David}. We can append the sampled token to the original sequence and input the new sequence (\texttt{Who directed Fight Club? David}) back into the model. The model returns a distribution over tokens which we can again sample from. Perhaps we get \texttt{Fincher}. We can then append that token to get \texttt{Who directed Fight Club? David Fincher}. This iterative process is how text is generated.

\section{Evaluation and Credence Attribution}

In this section, we characterise the emerging science of LLM evaluation, explain the role of LLM credence attribution within LLM evaluations, and distinguish three experimental techniques for assessing what credences LLMs have.

Model evaluation is a widespread practice in machine learning that aims to assess the performance of machine learning models at particular tasks. Central to model evaluation are \textit{benchmarks}.\footnote{See \citet{bender2021dangers}, \citet{denton2020bringing}, and \citet{weidinger2023sociotechnical} for sociotechnical critiques of benchmarking practices in machine learning.} For example, the General Language Understanding Evaluation (GLUE) benchmark provides training and test datasets alongside performance metrics for nine natural language understanding tasks including sentiment analysis and sentence classification, allowing for comparisons between models \citep{wang2019glue}. Similarly, the ImageNet Large Scale Visual Recognition Challenge provides a training and test dataset, and a performance metric, which serve as a common standard for evaluating image-based object recognition models (\citeauthor{russakovsky2015imagenet}, \citeyear{russakovsky2015imagenet}; see also \citeauthor{deng2009imagenet}, \citeyear{deng2009imagenet}).

Evaluating LLMs is more complex than evaluating \textit{narrow} models that are trained to perform a particular task such as image classification. LLMs are trained with a generic objective -- next token prediction -- that given broad training data enables the model to acquire an open-ended set of capabilities \citep{bommasani2021opportunities, chang2023survey}. The added challenge of LLM evaluation is identifying \textit{what} capabilities the model has acquired in training, alongside the standard task of benchmarking model performance on tasks that assess capabilities which the model is known to have. Furthermore, because LLMs input and output natural language, scientists have many degrees of freedom in terms of how model capabilities are elicited and assessed. For example, Jason \citet{wei2022chain} found that \textit{chain-of-thought} prompting, in which the model is prompted to reason through a problem `step-by-step,' substantially improves performance at arithmetic, commonsense, and symbolic reasoning tasks. Thus, LLM evaluations often centre on question-answering tasks that allow for particular kinds of answers. For example, an evaluation suite for logical reasoning might ask the LLM to assess whether exemplar inferences are valid or invalid \citep{saparov2024testing}, and may allow one-shot or chain-of-thought answering.

Increasingly, scientists purport to assess not only LLM answers to questions in evaluation suites, but also LLM confidence in the answers given. The idea is that rather than assessing whether or not the LLM correctly determines, for example, the validity of inferences, we can also assess the LLM's credence, or its degree of confidence, in its assessments of the validity of particular inferences. 

\begin{quote}
    [LLMs] do not produce a single answer; rather, they produce a probability distribution over the possible answers. This distribution can provide further insight into their processing. [...] [T]he probability assigned to the top answers, relative to the others, can be interpreted as a kind of confidence measure \citep[16]{dasgupta2022language}.
    
    [T]wo LLMs may achieve identical accuracy scores but exhibit different levels of uncertainty regarding the question [...] Consequently, it is necessary to incorporate uncertainty into the evaluation process to achieve a more comprehensive assessment \citep[2]{ye2024benchmarking}.
\end{quote}

LLM credence attributions are based on measurement techniques that seek to capture how confident an LLM is in a given proposition \citep{geng2023survey}. Techniques for measuring confidence in LLMs fall into three main buckets.\footnote{We bracket a fourth class of exploratory techniques that use interpretability methods to assess whether the LLM has any kind of internal representation corresponding to an assessment of the truth or falsity of a given proposition. \citet[][4-5]{geng2023survey} provide an overview. These techniques are discussed in Section \ref{Metaphysical Anti-Realism} in relation to the metaphysics of LLM credences.}

The first class of techniques involves prompting the LLM to generate a natural language report of its confidence in a given proposition \citep{xiong2023can, lin2022teaching, kadavath2022language}.\footnote{The technique described is the simplest of several report-based methods for eliciting confidence judgements \citep{xiong2023can}. For example, another elaborate method utilises chain-of-thought reasoning by prompting the LLM to reason through a problem ‘step-by-step' before providing its answer alongside a confidence judgement \citep[c.f.][]{wei2022chain}.} To illustrate: Suppose the LLM is prompted with: \texttt{How confident are you that Fincher directed Fight Club?}. The idea is that if the LLM outputs, for example, \texttt{75\%}, then the LLM has a credence of 75\% in the proposition that $\ulcorner$Fincher directed Fight Club$\urcorner$.

The second class of techniques uses consistency-based estimation to infer the LLM's credence in a proposition \citep{manakul2023selfcheckgpt}. To be sure, one problem with inferring LLM credences from reported confidence judgements is that LLMs are liable to return different confidence judgements across multiple trials given the stochastic process by which LLMs generate text. Consistency-based estimation tries to address this issue by inferring the LLM's degree of confidence in a proposition from the degree of variation exhibited in the LLM's answers when the LLM is repeatedly asked about that proposition. To illustrate: If the LLM returns \texttt{David Fincher} in $95$ out of $100$ independent trials in which it is prompted to answer the question \texttt{Who directed Fight Club?}, we can infer that the LLM has high confidence of around $95\%$ in the proposition that $\ulcorner$Fincher directed Fight Club$\urcorner$. On the other hand, if the LLM returns \texttt{David Fincher} in $40$ instances, \texttt{Quentin Tarantino} in $25$ instances, and \texttt{Sofia Coppola} in $35$ instances, then we can infer that the LLM has only moderate confidence of around $40\%$ in the proposition that $\ulcorner$Fincher directed Fight Club$\urcorner$.\label{Measuring}

The third class of techniques involves deriving a confidence judgement from the LLM's output probabilities conditional on a a prompt which instructs the LLM to affirm or deny a proposition (\citeauthor{lin2022teaching}, \citeyear{lin2022teaching}; \citeauthor{kuhn2023semantic}, \citeyear{kuhn2023semantic}; \citeauthor{kadavath2022language}, \citeyear{kadavath2022language}; see also \citeauthor{dasgupta2022language}, \citeyear{dasgupta2022language}, 16). To illustrate: Suppose the LLM is prompted with: \texttt{Fincher directed Fight Club. True or False?}. The idea is that if the LLM's output probability for \texttt{True} is 75\%, then the LLM has a credence of 75\% in the proposition that $\ulcorner$Fincher directed Fight Club$\urcorner$.

\section{Semantics}

How should we interpret LLM credence attributions? Are scientists quite literally ascribing degrees of confidence to LLMs; or are LLM credence attributions a stylised shorthand that readily translates into non-mentalistic language? In this section, we argue that LLM credence attributions should (at least in general) be interpreted literally; that is, as expressing truth-apt beliefs on the part of scientists that purport to describe facts about LLM credences. 

First, there is a presumptive case for interpreting LLM credence attributions literally. LLM credence attributions \textit{appear} to describe facts about LLM credences. Accordingly, there is a defeasible reason to suppose that LLM credences are literal ascriptions of credences to LLMs. The presumptive case is reinforced by the fact that scientists evidence LLM credence attributions with experimental measurement techniques. That scientists give empirical justifications for LLM credence attributions indicates a background commitment to the claim that there is a fact of the matter about what the LLM's credences are. 

The presumptive case can be strengthened with an argument from elimination. The two most obvious non-literal interpretations of LLM credence attributions are implausible. On one hand, it may be that LLM credence attributions are shorthand for statements about the token succession probabilities given by LLMs conditional on token sequences. This interpretation would be plausible if attributions of credences to LLMs were restricted to propositions of the form $\ulcorner$$t$ succeeds $s$$\urcorner$ for tokens $t$ $\in$ $\mathcal{T}$ and token sequences $s$ $\in$ $\mathcal{T}^N$. Then utterances such that the LLM has such-and-such confidence in $\ulcorner$$t$ succeeds $s$$\urcorner$ could be interpreted as a stylised shorthand for the non-mentalistic claim that the LLM evaluates $p(t|s)$ as such-and-such. In practice LLMs are attributed credences in a broader class of propositions than those pertaining to token succession \citep[c.f.][]{kuhn2023semantic, shanahan2023role}. So, LLM credence attributions do not readily translate to non-mentalistic claims about token succession probabilities.

On the other hand, it may be that LLM credence attributions are shorthand for claims about probabilities that can be \textit{elicited} from LLMs, but which do not reflect credences on the part of LLMs. Techniques for measuring LLM credences often originate in engineering programs with practical aims. For example, one approach to tackling LLM-generated misinformation involves fine-tuning LLMs to supplement assertions with calibrated probability estimates. These probability estimates can then be used as indicators for how much evidential weight to assign the LLM's assertions \citep{kadavath2022language, lin2022teaching, manakul2023selfcheckgpt}.\footnote{Here an LLM's probability estimates are calibrated if, for example, 80\% of the propositions to which the LLM assigns a probability of 80\% are true.} If what matters is getting LLMs to provide calibrated probability estimates, then it is sufficient for LLMs to return calibrated probability estimates via some observable signal. It does not matter whether the probability estimates reflect underlying credences. Perhaps, then, experimental techniques used by scientists are not \textit{measuring} credences, but rather \textit{eliciting} probabilities that can be engineered to satisfy practically useful properties such as calibration. 

This latter reading of LLM credence attributions fails to account for the stated motivations of scientists developing measurement techniques for LLM credences. Typically, their stated motivation is training LLMs to be \textit{honest}, in both the objective sense that the LLM `should give accurate information' and the subjective sense of being `honest about itself and its internal state' (\citeauthor{askell2021general}, \citeyear{askell2021general}, 5; see also \citeauthor{bai2022training}, \citeyear{bai2022training}; \citeauthor{evans2021truthful}, \citeyear{evans2021truthful}). Consider,

\begin{quote}
    We would eventually like to train AI systems that are honest, which requires that these systems accurately and faithfully evaluate their level of confidence in their own knowledge and reasoning 
    (\citeauthor{kadavath2022language}, \citeyear{kadavath2022language}, 2; see also \citeauthor{lin2022teaching}, \citeyear{lin2022teaching}, 2; \citeauthor{yang2023alignment}, \citeyear{yang2023alignment}).
\end{quote}

The goal of training honest systems requires scientists to register the distinction between \textit{measuring} LLM credences and merely \textit{eliciting} calibrated probabilities from LLMs. Indeed they do. Stephanie \citet[2]{lin2022teaching}, for example, claim that `[c]alibration is compatible with a certain kind of dishonesty, because a model could be calibrated by simply imitating a calibrated individual.' Similarly, Amanda \citet[5]{askell2021general} claim that `it is not sufficient for [LLMs] to simply imitate the responses expected from a seemingly humble and honest expert.' That the possibility of LLMs merely imitating a calibrated individual is acknowledged shows a background commitment to LLMs having credences. The goal is not merely that of eliciting calibrated probabilities from LLMs, but rather that of measuring the LLM's confidence in particular propositions.

These arguments provide a strong case for accepting a literal interpretation of LLM credences. We leave open that some LLM credence attributions advanced by some scientists may be intended non-literally. We nevertheless maintain that LLM credence attributions at least in general express truth-apt beliefs on the part of scientists that purport to describe facts about LLM credences. 

\section{Metaphysics}\label{Metaphysical Anti-Realism}

If LLM credence attributions are correctly interpreted literally, the next obvious question is whether LLM credences exist. For if LLMs lack credences, then LLM credence attributions are all subject to a presupposition failure, and LLM credence attributions as a class of statements are subject to wholesale error. We first clarify the claim that LLM credences exist and then argue that, while this claim is somewhat plausible, the evidence for its acceptance is inconclusive.

\subsection{The Existence Claim}

What does it mean to say that LLM credences \textit{exist} or that LLMs \textit{have} credences? Traditionally, credences admit two rival interpretations that correspond to the distinction between \textit{mentalist} and \textit{behaviourist} interpretations of decision-theory (\citeauthor{dietrich2016mentalism}, \citeyear{dietrich2016mentalism}; \citeauthor{okasha2016interpretation}, \citeyear{okasha2016interpretation}; see also \citeauthor{hansson1988risk}, \citeyear{hansson1988risk}; \citeauthor{bermudez2009decision}, \citeyear{bermudez2009decision}). First, mentalists think that credences are psychologically real mental states that, together with utilities, explain an agent's behaviour. On this view, credences and utilities are quantitative extensions of belief-desire psychology, where actions are causally explainable with reference to gradational as opposed to categorical belief-like states (\citeauthor{lewis1974radical}, \citeyear{lewis1974radical}, 337; see also \citeauthor{hampton1994failure}, \citeyear{hampton1994failure}; \citeauthor{okasha2016interpretation}, \citeyear{okasha2016interpretation}). Second, behaviourists think that credences are formal constructs that alongside utilities are derivable from an agent’s revealed preferences. This latter view is widely endorsed in microeconomics where, in the tradition of \citet{savage1972foundations}, the idea is that an agent can be represented as an expected utility maximiser relative to a credence function and a family of utility functions so long as their revealed preferences satisfy certain rationality constraints.\footnote{\citeauthor{savage1972foundations}'s (\citeyear{savage1972foundations}) decision theory takes as basic a set of outcomes and a set of states of the world, such that acts that an agent must choose between can be understood as functions from states to outcomes. The agent is assumed to have preferences over acts, and Savage’s theorem says that an agent is representable as an expected utility maximiser relative to a unique credence function and a utility function that is unique up to positive linear transformations just in case the preferences satisfy certain axioms including transitivity, reflexivity, and the sure-thing principle. Savage follows in the tradition of \citet{von2007theory} who proved an analogous result which takes objective probabilities as primitive and says that provided an agent’s preferences adhere to certain rationality constraints, then that agent can be represented as an expected utility maximiser relative to a utility function that is unique up to positive linear transformations and the objective probability function.} 

The claim that LLM credences exist can be clarified in light of the dispute between mentalism and behaviourism. First, the claim that LLM credences exist should be read along mentalist lines. On the one hand, were LLM scientists attributing LLMs credences in the behaviourist sense, then the evidence provided in support of particular LLM credence attributions would be consistent choice behaviour on the part of the LLM and not, \textit{inter alia}, reported confidence judgements and the relative frequencies of answers over multiple independent trials. On the other hand, LLM credence attributions are typically motivated on grounds that LLM credences provide `insight into [LLM] processing' \citep[16]{dasgupta2022language}. Yet no such insight could be provided if the LLM credences at issue were causally inert formal constructs derived from LLM choice behaviour. 

Second, LLM credences are supposedly the same \textit{kind} of mental state as human credences. The idea is not that LLMs can have some other kind of mental state (say, \textit{schmedences}) that are in certain respects similar to credences. Rather, LLM credences are thought to play the same kind of causal explanatory role in LLM cognition as credences play in human cognition. Consider,

\begin{quote}
    LLMs may achieve identical accuracy scores but exhibit different levels of uncertainty regarding the question. This is analogous to students taking exams [...] where two students may select the same answer but actually possess distinct degrees of uncertainty or comprehension about the question \citep[1]{ye2024benchmarking}.
\end{quote}

Last, the idea that LLMs and humans both have credences allows for LLMs and humans to realise credences differently. Humans realise credences in a neurobiological substrate whereas LLM credences are realised in a digital substate. Minimally, for this picture to work, we need to accept the metaphysical thesis that credences are \textit{multiply realisable} such that credences as mental states can be realised by different physical kinds \citep[c.f.][]{putnam1967psychological}. The simplest way to do this is to endorse \textit{functionalism}, the idea that mental states are functional states of a system; and second, \textit{computationalism}, the idea that the specific kind of functional states that realise mental states are computational states.\footnote{See \citet[105ffff]{fodor2000mind} on the distinction between computationalism and functionalism. Endorsing computationalism and functionalism about credences does not entail a commitment to computationalism and functionalism about mental states in general. Hence one can consistently say that LLMs have credences while denying the possibility that LLMs have phenomenal states \citep[c.f.][13-14]{butlin2023consciousness}. Nor is this restriction \textit{ad hoc}. Indeed, \citeauthor{putnam1967psychological}'s (\citeyear{putnam1967psychological}) original treatment of multiple realisability was in the context of machine functionalism on which computationalism and functionalism hold for all mental states. But \citeauthor{fodor1975language}'s (\citeyear{fodor1975language}) later representational theory of mind applies computationalism to propositional attitudes only \citep[see also][]{fodor1983representations, fodor1987psychosemantics, fodor1992theory, fodor1995elm, fodor2008lot}.}

\subsection{Inference to the Best Explanation}

We now assess the case for the existence of LLM credences. We argue that the existence of LLM credences is at least plausible, but that current evidence is inconclusive. What, then, is the case for accepting that LLM credences exist?

One option is to say that LLMs report having credences in propositions, and these reports provide a defeasible reason to suppose that LLMs have credences. This argument is suspect because LLMs are trained to reliably mimic human language via mechanisms that differ from the cognitive mechanisms that allow for language in humans \citep[][4]{butlin2023consciousness}. Hence LLM reports of credences may be subject to a debunking explanation that undercuts the evidential relationship that ordinarily obtains between reports of credences and credences. Whether and under what conditions LLM reports of internal states provide evidence for internal states is not yet well understood \citep{perez2023towards}. Accordingly, we set aside LLM self-reports of credences as evidence for the existence of LLM credences, but nevertheless leave open the possibility that LLM report data may under the right circumstances carry evidential weight.

What seems to be an altogether more promising option is to argue for the existence of LLM credences via an inference to the best explanation.\footnote{This argument is a close parallel to the no-miracles argument for scientific realism, on which it would be a \textit{miracle} if our best scientific theories were predicatively successful to the extent that they are if their claims about unobservable entities were not at least approximately true. Hence we can infer the approximate truth of claims about unobservables on grounds that their existence best explains the predictive success of our best theories \citep{putnam1975mathematics}.} The idea of an inference to the best explanation is to infer that LLM credences \textit{exist} on grounds that LLM credences \textit{best explain} certain capabilities exhibited by LLMs. Consider, for example, the ability of LLMs to perform spatial reasoning. Plausibly, what best explains the ability of LLMs to, for example, navigate a simulated environment through text-based commands such as ‘move left' and ‘move right', and then provide a detailed description of the spacial features of that environment based on textual feedback cues \citep[51]{bubeck2023sparks}, is that the LLM has internal states corresponding to degrees of confidence \textit{about} the simulated environment. If so, then we can then infer the \textit{existence} of LLM credences on grounds that LLM credences best explain LLM spatial reasoning.

This argument is too quick. LLM credences are a \textit{possible} explanation of capabilities like spatial reasoning, but it is a stretch to say that credences are the best or even a plausible explanation of these capabilities. It could also be that such capabilities are explained by the LLM registering statistical patterns in language that allow for successful next word prediction on (for example) spatial reasoning tasks, and so more work is required to establish the plausibility of credences as an explanation for the relevant capabilities \citep{bender2021dangers}.

To be sure, there is some evidence that points towards credences as an explanation of LLM capabilities. LLMs have internal representations of the truth-values of propositions. In particular, it is possible to train a separate neural network (called a \textit{probe}) to predict better than chance the truth-values of propositions from LLM hidden layer activations that obtain when LLMs process token sequences that express those propositions \citep{azaria2023internal, herrmann2024standards}. That LLM hidden layer activations realise statistical patterns that allow for better-than-chance prediction of truth-values suggests that the internal states of LLMs contain information about the truth-values of propositions. The presence of such information motivates the plausibility of LLMs having propositional attitudes \textit{of some kind}. But these attitudes need not be belief-like. (It would, for example, be possible to predict the truth-values of propositions better-than-chance if LLMs had desires about the relevant propositions where the strength of the LLM's desires was correlated with the truth-values of the relevant propositions.) Furthermore, even if LLMs represent the truth-values of propositions in a belief-like way, it is not clear that the particular kind of representation is credence-like; namely, pairs of propositions and associated scalar quantities corresponding to degrees of confidence. 

Overall, current evidence for the existence of LLM credences is inconclusive. There is clear potential to argue for the existence of LLM credences on grounds that LLM credences best explain certain LLM capabilities. But at present LLM credences are at best a plausible explanation of the relevant LLM capabilities, and the commitment that LLMs have credences is metaphysically suspect. This point underwrites a non-trivial concern that LLM credence attributions are systematically false because all such claims are subject to a presupposition error.

\section{Epistemology}\label{Epistemic Anti-Realism} 

Even if LLMs have credences, it is possible that techniques such as reported confidence, consistency-based estimation, and measuring output probabilities, fail to track facts about LLM credences. If so, then attributions of credences to LLMs made on the basis of these techniques are not even generally true. 

We discuss several challenges to reported confidence, consistency-based estimation, and output probabilities, as indicators for LLM credences. We argue, first, that reported confidence is not a reliable indicator of LLM credences. Second, we argue that under certain conditions consistency-based estimation may be a reliable indicator of LLM credences, but at present we lack a clear mechanism which guarantees that signals about the LLM's credences reliably show up in the distribution of answers over multiple independent trials. Third, we argue that it is at best unclear how to infer LLM credences from output probabilities given that output probabilities apply to syntactic units -- \textit{tokens} -- as opposed to propositions. We conclude that \textit{even if} LLMs have credences, we may well be in a sceptical scenario in which techniques for evidencing credences are unreliable.

\subsection{Stochasticity and Reported Confidence} \label{epistem-report}

We start with the most straightforward method for assessing LLM credences; namely, prompting the LLM to report its confidence in a proposition \citep{xiong2023can, lin2022teaching, kadavath2022language}. Reported confidence judgements from LLMs cannot \textit{directly} indicate LLM credences in the sense that an LLM has a credence $c$ in $X$ just in case the LLM reports a credence $c$ in $X$. 

For reported confidence judgements to evidence LLM credences directly, we need a mechanism to ensure that identifying signals for the LLM's credences show up in the LLM's reported confidence judgements. However, LLM reported confidence judgements are generated by a stochastic sampling process. So, any mechanism linking the LLM's credences to its reported confidence judgements is mediated by a stochastic process such that whatever credence the LLM reports is one of several judgements that could have been reported (each with an associated probability). Hence LLM reported confidence judgements cannot \textit{directly} indicate what credences LLMs have because \textit{even if} an LLMs has credences, it is a matter of chance whether the LLM correctly reports its credence.\footnote{The argument here also generalises to more complex techniques for eliciting reported confidence judgements from LLMs such as those which employ chain-of-thought reasoning \citep{geng2023survey}. For any such technique, the confidence judgement reported is one of several that could be reported, each with an associated probability.}

\subsection{Distorting Factors and Consistency-Based Estimation} \label{epistem-consistency}

Perhaps we can infer LLM credences \textit{indirectly} from the distribution of reported credences over many independent trials. For example, given the prompt \texttt{How confident are you that Fincher directed Fight Club?}, we could plausibly infer that the LLM is around 95\% sure that $\ulcorner$Fincher directed Fight Club$\urcorner$ if over 100 trials the LLM returns \texttt{95\%} as the most frequent response, and the vast majority of the LLM's responses are \texttt{93\%}, \texttt{94\%}, \texttt{95\%}, \texttt{96\%}, and \texttt{97\%}. Plausibly, a similar picture holds for humans, in which having a credence $c$ in $X$ may dispose a human to report confidence levels in the ballpark of $c$, but where any particular reported confidence is an unreliable indicator of the credence, $c$.

This approach to inferring LLM credences from reported confidence judgements is a form of consistency-based estimation \citep{manakul2023selfcheckgpt}. We could similarly prompt the LLM 100 times with \texttt{Did Fincher direct Fight Club?}. Then if the LLM responds \texttt{Yes} in a large majority of cases, we can supposedly infer that the LLM has high confidence that $\ulcorner$Fincher directed Fight Club$\urcorner$. The principle in both cases is the same: Informative signals about LLM credences supposedly show up in the distribution of the LLM's responses conditional on an appropriately selected prompt over multiple independent trials.

We argue that the distribution of answers reliably signals the LLM's credences only if certain conditions obtain. Distorting factors that have the potential to skew the distribution including non-standard temperature values and non-standard sampling methods need to be avoided. Even then, we lack a positive account of the mechanism by which signals about the LLM's credences reliably show up in the distribution of answers over multiple independent trials. 

\subsubsection{Temperature}

First, the relative frequencies of answers given by the LLM in response to a prompt over multiple independent trials depend in part on the LLM's output probabilities for tokens conditional on the initial input sequence.\footnote{This is a simplification that holds for one-token responses. For multiple-token responses we consider the initial output probabilities and the output probabilities and subsequent inputs constructed by iteratively appending sampled tokens to prior inputs. For an input sequence $s$ $=$ $(t_1,...,t_n)$ $\in$ $\mathcal{T}^N$ , the probability of a single-token response $t' \in \mathcal{T} $ is the model's output probability for $t'$ given input $s$, i.e. $p(t'|s)$. To get the probability for multi-token responses we multiply out the probabilities for single-token responses. For example, a two token response $(t',t'') \in \mathcal{T}^2$ is $p(t,t'|s) = p(t|s) \times p(t''|(s,t'))$ \citep[][3]{kuhn2023semantic}.} Given random sampling, the relative frequencies of (single-token) responses over independent trials will in the limit converge on the output probabilities for tokens.

What probabilities the LLM outputs conditional on a given input also depends on a parameter called \textit{temperature}. This parameter is set by the user and controls the smoothness of the probability distribution. To explain: Output probabilities are generated in the final layer of the neural network. The penultimate layer computes \textit{logits} for each token. Logits are real-numbers that can be positive or negative and which indicate how confident the LLM is that each token succeeds the input sequence. Logits are converted into probabilities (that is, real-numbers between 0 and 1 that sum to 1) in the final layer via \textit{softmax normalization}. Logit $x_k$ for token $t_k$ is converted to probability $p_k$ as follows:

\[
    p_k = \frac{e^{\frac{x_k}{T}}}{\sum_{i=1}^{N}e^{\frac{x_i}{T}}}
\]

The top half of the fraction ensures that the output is positive because $f(x)$$=$$e^x$ is positive in all its arguments, and also preserves the order of the logits because $f(x)$$=$$e^x$ is strictly increasing. The bottom half of the fraction ensures that the outputs sum to one. Hence softmax normalisation converts logits into probabilities in a way that ensures that tokens with greater logit values receive a greater share of the probability mass. The temperature $T$ determines the smoothness of the distribution. First, take the case where $T$ $\in$ $(0,1)$. As $T$ $\rightarrow$ $0$, the effect is to increase the distances between logits, resulting in more extreme differences in probabilities. So, comparatively higher probability tokens receive a greater fraction of the total probability mass. Second, take the case where $T$ $\in$ [1,$\infty$). Here as $T$ $\rightarrow$ $\infty$, the effect is to decrease the distance between logits, such that in the limit the distribution converges on a uniform distribution that assigns equal probability to each token. Hence for higher temperatures comparatively higher probability tokens receive a smaller fraction of the total probability mass.

The practical significance of temperature is that temperature controls the \textit{creativity} or \textit{predictability} of text generated when sampling from the LLM. Consider the token sequence \texttt{The cat sat on the}. When $T$ is low, the bulk of the probability mass will be assigned to comparatively higher probability tokens such as \texttt{mat} and \texttt{chair}, such that sampling from the LLM will typically yield predictable non-creative completions. When $T$ is high, the probability mass will be more uniformly distributed across tokens rendering it more probable that comparatively less likely tokens such as \texttt{train} or \texttt{pear} will be sampled. Hence higher temperature yields less predictable and thus more creative completions. 

It is hard to see how the distribution of answers given by the LLM could reliably indicate the LLM's credences given that the observed distribution depends on the user's choice of temperature - an exogenous factor. Holding fixed the input sequence, each temperature value yields a unique set of output probabilities, and these output probabilities determine the relative frequencies of responses over multiple independent trials. Presumably, if LLMs have credences, then those credences are endogenous to the LLM in the sense of being encoded in the LLM's model weights. Because the distribution of responses depends in part on the exogenous factor of the user's choice of temperature, information about the LLM's credences in the distribution of responses is liable to distortion. Hence the distribution of responses over multiple independent trials cannot reliably signal the LLM's credences as the chosen temperature is a distorting factor.

One plausible response is to say that there exists a particular \textit{non-arbitrary} temperature value. The idea is that output probabilities indexed to the non-arbitrary temperature value generate answer distributions that are indicative of the LLM's credences. But output probabilities indexed to other temperature values generate distorted distributions. The most obvious candidate for a non-arbitrary temperature value is whatever value was used when training the LLM (typically $T=1$). Output probabilities indexed to this temperature value reflect the LLM's probabilistic estimates with respect to token succession arrived at via a trial-and-error process in which the model's parameters are iteratively updated to correct for erroneous predictions. In this respect output probabilities indexed to the temperature value used in training are non-arbitrary. Hence LLM answer distributions need not be excluded as as a basis for inferring LLM credences on the basis of temperature-dependence, provided we accept the assumption that the relevant distributions are generated with the non-arbitrary temperature.

\subsubsection{Sampling Methodology}

Nevertheless, the problem does not obviously resolve even if we grant the existence of a non-arbitrary temperature. Consistency-based estimation requires sampling responses from the LLM. But the user's choice of sampling methodology impacts the distribution of responses in much the same way as temperature. 

To explain: Suppose we input the sequence: \texttt{How confident are you that Fincher directed Fight Club?}. First, the LLM computes a probability distribution over tokens conditional on the input. A token is sampled from the distribution. Suppose we get \texttt{99}. The sampled token is appended to the input. The resultant sequence (\texttt{How confident are you that Fincher directed Fight Club? 99}) is inputted into the LLM. The LLM computes a distribution, and we sample another token. Suppose we get \texttt{\%}. Hence we get a 99\% confidence judgement in the proposition $\ulcorner$Fincher directed Fight Club$\urcorner$. Two common sampling methods are top-$p$ and top-$k$ sampling. Top-$p$ samples a token from the smallest set of tokens whose cumulative probability exceeds some threshold, $p$. The successor token is selected from this set via a weighted lottery taking into account the probabilities assigned to each token. In contrast, top-$k$ sampling samples from the $k$ tokens with the highest probability, and again the sampling is weighted by the probabilities that the LLM assigns to each token. The distribution of answers that we observe across a given number of trials depends in part on the LLM's output probabilities for each token, but it also depends on the sampling methodology used. For example, whether top-$p$ or top-$k$ sampling is chosen, and the chosen value for $p$ and $k$ respectively. This latter dependency is problematic. For we could in effect force the LLM to exhibit greater or lesser consistency in its answers through our chosen sampling methodology. For example, using top-$p$ sampling with a value of $p$ close to $0$ would all but guarantee a highly consistent set of answers; whereas top-$p$ sampling with a value of $p$ close to $1$ would allow for more variability in the LLM's answers. The concern is that the degree of consistency exhibited in the LLM's answers is more a reflection of our chosen sampling methodology than a reflection of the LLM's credences. 

In response, perhaps, as with temperature, there exists some non-arbitrary way of sampling from the LLM. We suggest that employing top-\textit{p} sampling with $p=1$ is a non-distorting sampling method. For this sampling method amounts to a weighted lottery over the LLM's outut probabilities where the weights on each token are equal to the LLM's output probabilities for each token. Top-\textit{p} sampling with $p=1$ is, accordingly, indicative of the probability distribution over tokens learned by the LLM in training. 

\subsubsection{Summary}

The user's choice of temperature and sampling methodology have the potential to distort information about the LLM's credences contained in the distribution of answers over multiple independent trials. We have suggested that the distortion issue can be overcome by using a temperature of $1$ alongside top-\textit{p} sampling where $p=1$. Even so, while the distortion issue is resolvable, we lack a positive rationale for accepting an evidential relationship between the distribution of answers and the LLM's credences. We leave open the possibility that a rationale can be provided here, but in the meantime suggest caution in attributing LLMs credences based on the distribution of answers over multiple independent trials.

\subsection{Syntax, Semantics, and Output Probabilities} \label{syntax_and_semantics}

We now argue that LLM credences cannot obviously be inferred from output probabilities. The problem is that output probabilities are indexed to \textit{tokens} but credences are indexed to \textit{propositions} \citep{kuhn2023semantic}. Hence it is not clear how we are supposed to infer the LLM's credences from output probabilities. We need a principle which specifies the evidential relationship between output probabilities and credences. What would such a principle look like? Consider,

\begin{quote}
    \textit{The Simple Bridge Principle:} The LLM has a credence $cr(X)$ in $X$ just in case the LLM assigns a probability $cr(X)$ to the \texttt{Yes} token given an input prompt $s_X$ for the LLM to affirm or deny $X$.
\end{quote}

To illustrate: Suppose the LLM is prompted with: \texttt{Did Fincher direct Fight Club?}. Then, on the simple bridge principle, the LLM's credence in the proposition that $\ulcorner$Fincher directed Fight Club$\urcorner$ is equal to the probability that the LLM assigns to the \texttt{Yes} token conditional on the input \texttt{Did Fincher direct Fight Club?} \citep{kadavath2022language, tian2023just}. The rationale is that the LLM's credence in the affirmative \texttt{Yes} token is indicative of the LLM's degree of confidence in the proposition that $\ulcorner$Fincher directed Fight Club$\urcorner$.

The simple bridge principle fails. Given the input \texttt{Did Fincher direct Fight Club?}, the LLM can affirm the proposition using one of several semantically equivalent tokens. For example, the tokens \texttt{Yes}, \texttt{yes}, \texttt{sure}, and \texttt{absolutely} all represent affirmative answers to the question. The idea that the LLM's credence in the proposition that $\ulcorner$Fincher directed Fight Club$\urcorner$ is indexed to the output probability assigned to the \texttt{Yes} token is implausible given the existence of other token responses that also represent affirmative answers. 

One patch is to partition the tokens in the LLM's vocabulary into semantically equivalent subsets, and take the sum of probabilities for all affirmative tokens as the LLM's credence in the relevant proposition \citep{kuhn2023semantic}.

\begin{quote}
    \textit{The Additive Bridge Principle:} The LLM has a credence $cr(X)$ in $X$ just in case $cr(X)$ is equal to the sum of the LLM's output probabilities for the \texttt{Yes} token and its semantic equivalents given an input prompt $s_X$ for the LLM to affirm or deny $X$.
\end{quote}

Formally, let $E$ $\subseteq$ $\mathcal{T}^2$ be an equivalence relation such that for two tokens $t,t'$ $\in$ $\mathcal{T}$, $(t,t')$ $\in$ $E$ just in case $t$ and $t'$ are semantically equivalent. Put $\mathcal{C}_\texttt{Yes}$ as the equivalence class of tokens that are semantically equivalent to the \texttt{Yes} token:

\[
   \mathcal{C}_\texttt{Yes} = \{t \in \mathcal{T} | t \sim_{E}\texttt{Yes}\}
\]

The additive bridge principle holds that the LLM's credence $cr(X)$ in $X$ is equal to the sum of the LLM's output probabilities for all tokens $t$ $\in$ $\mathcal{C}_\texttt{YES}$ conditional on the prompt $s_X$ for the LLM to affirm or deny $X$. That is,

\[
cr(X) = \sum_{t \in \mathcal{C}_\texttt{Yes}} p(t | s_X)
\]

The additive principle captures the rationale behind the simple principle. In particular, that the probability assigned by the LLM to an affirmative completion indicates its credence in the relevant proposition. But it avoids the charge of arbitrarily privileging one affirmative token over semantically equivalent alternatives. So, the additive principle trumps the simple principle, all else equal.

The additive principle does not work either. Given the input \texttt{Did Fincher direct Fight Club?}, LLMs can and do assign non-zero probabilities to tokens that represent neither affirmations nor denials of the proposition in question. For example, \texttt{triangle} and \texttt{Basingstoke}. Hence the sum of probabilities assigned to \texttt{Yes} and \texttt{No} and their semantic equivalents will not in practice sum to 1. 

One option is to say that only the output probabilities corresponding to \texttt{Yes} and \texttt{No} and their semantic equivalents are \textit{evidentially relevant} to the LLM's credence in the proposition that $\ulcorner$Fincher directed Fight Club$\urcorner$. What makes these probabilities evidentially relevant is the semantic content of the tokens to which they are assigned. In particular, \texttt{Yes} and \texttt{No} and their semantic equivalents correspond to legitimate epistemic attitudes that the LLM could hold towards the proposition; namely, the attitudes of affirmation and denial. In contrast, output probabilities assigned to spurious tokens such as \texttt{triangle} are \textit{evidentially irrelevant} to the LLM's credence. For these probabilities are indexed to tokens which fail to correspond to legitimate epistemic attitudes. This distinction, if robust, motivates a bridge principle that considers only the probability mass assigned to the \texttt{Yes} and \texttt{No} tokens and their semantic equivalents. Consider,

\begin{quote}
    \textit{The Normalised Bridge Principle:} The LLM has a credence $cr(X)$ in $X$ just in case $cr(X)$ is equal to the sum of the LLM's output probability for the \texttt{Yes} token and its semantic equivalents divided by the sum of the \texttt{Yes} and the \texttt{No} tokens and their semantic equivalents given an input prompt $s_X$ for the LLM to affirm or deny $X$.
\end{quote}

\[
cr(X) = \frac{\sum_{t \in \mathcal{C}_\texttt{Yes}} p(t | s_X)}{\sum_{t \in \mathcal{C}_\texttt{Yes}} p(t | s_X) + \sum_{t \in \mathcal{C}_\texttt{No}} p(t | s_X)}
\]

The normalised bridge principle fails. The rationale for this principle is that output probabilities for affirmative and denial tokens are evidentially relevant to the LLM's credences given their semantic content, but output probabilities assigned to spurious tokens are evidentially irrelevant. But the affirmative, denial, and spurious tokens do not exhaust the LLM's vocabulary. We can distinguish between spurious tokens such as \texttt{triangle} and tokens which are neither affirmations nor denials but which indicate a legitimate epistemic attitude towards the proposition. These include, \textit{inter alia}, \texttt{unlikely}, \texttt{probably}, \texttt{perhaps}, and \texttt{unsure}. We cannot admit affirmative and denial tokens as evidentially relevant to the LLM's credences \textit{because of} their semantic content, but exclude tokens whose semantic content reflects other epistemic attitudes as evidentially irrelevant \textit{despite} their semantic content. Hence the normalised principle fails.

There may exist a more complex bridge principle that factors in output probabilities assigned to all tokens which correspond to legitimate epistemic attitudes. Yet such a principle would face the issue that output probabilities assigned to tokens and the epistemic attitudes denoted by the tokens give two separate scales for assessing the LLM's epistemic attitude and there is no obvious way to combine the two. It is at best unclear how to combine (for example) a 70\% output probability for \texttt{Yes} and a 12\% output probability for \texttt{probably}, among other pairs of output probabilities and legitimate epistemic attitudes, into a single scalar value representing the LLM's credence in a proposition.

Even if some bridge principle allows us to factor in the output probabilities assigned to all tokens which correspond to legitimate epistemic attitudes alongside the attitudes denoted by the relevant tokens, a second problem remains. Output probabilities are not invariant under semantically equivalent formulations of the same question \citep[c.f.][]{scherrer2024evaluating}. LLMs can and do give different output probabilities conditional on semantically equivalent but syntactically distinct prompts such as \texttt{Did Fincher direct Fight Club?} and \texttt{Is it true that Fincher directed Fight Club?}. Any principle that links the LLM's credence in $X$ to the LLM's output probabilities conditional on some input sequence $s_X$ for the LLM to affirm or deny $X$ arbitrarily privileges one of several semantically equivalent prompts for the LLM to affirm or deny $X$. This issue motivates the need for an even more complex bridge principle that accounts for semantic equivalence of the input sequences in addition to the issues identified above. It is at best unclear to us what this principle would be.\footnote{Plausibly, a variant of consistency-based estimation could be leveraged to address the problem that LLM output probabilities are not invariant under semantically equivalent formulations of the same question \citep[c.f.][]{scherrer2024evaluating}. But such a complex technique for estimating LLM credences would require independent theoretical motivation for why signals of the LLM's credences would reliably show up using this measurement technique.}\footnote{Our discussion has so far only addressed single token responses. The issues around semantic equivalance generalise to multiple-token responses, where multiple-token responses are calculated by multiplying out single-token responses \citep{kuhn2023semantic}. But the problem is further complicated in the multiple-token case because there are a countably infinite infinite number of semantically equivalent token sequences that amount to affirmations or denials of a proposition $X$ when multiple-token responses are admitted.}

\subsection{Interdependence}
We considered three sources of evidence about LLM credences: reported confidence, consistency-based estimation, and output probabilities. We argued that reported confidence is an unreliable indicator of LLM credences, and that it is at best unclear how to infer LLM credences from output probabilities. We further argued that consistency-based estimation could in principle signal the LLM's credences given an appropriate choice of temperature and sampling methodology, but that a positive mechanism is lacking which guarantees that signals about the LLM's credences reliably show up in the distribution of answers. 

While we have presented reported confidence, consistency-based estimation, and output probabilities separately, their limitations are interdependent. Holding fixed LLM model weights, the input sequence and the temperature determine output probabilities. Output probabilities and sampling methods (such as top-$p$ and top-$k$) determine the distribution from which token responses are sampled, and in turn the relative frequencies of different answers across multiple independent trials. Hence in all three cases endogenous signals of the LLM's credences are liable to distortion from exogenous factors such as the user's choice of temperature and sampling method. LLM scientists should avoid distorting factors when attributing credences to LLMs on the basis of any of these techniques. 

\section{Conclusion} 

Scientists often attribute credences to LLMs when evaluating LLM capabilities. In this paper, we asked three questions about LLM credence attribution. First, are LLM credence attributions literal ascriptions of degrees of belief or confidence to LLMs? Second, do LLMs have credences? Third, even if LLMs have credences, are the experimental techniques used to assess LLM credences reliable? Our analysis suggests that LLM credence attributions are (at least in general) intended by scientists as literal ascriptions of credences to LLMs. Evidence for the existence of LLM credences is at best inconclusive, and even if LLMs have credences, there are non-trivial questions about the reliability of the techniques used to evidence LLM credences -- namely, reported confidence, consistency-based estimation, and output probabilities. It is a distinct possibility that \textit{even if} LLMs have credences, LLM credence attributions made on the basis of existing experimental techniques are not even generally true.

\bibliography{references}

\end{document}